\documentclass[10pt,twocolumn,letterpaper]{article}

\usepackage[pagenumbers]{cvpr} 

\definecolor{cvprblue}{rgb}{0.21,0.49,0.74}
\usepackage[pagebackref,breaklinks,colorlinks,allcolors=cvprblue]{hyperref}
\usepackage{algorithm}
\usepackage{algorithmicx}
\usepackage{multirow}
\usepackage{graphicx}
\usepackage{amsmath,amssymb,amsthm}
\usepackage{adjustbox}
\usepackage{booktabs}       
\usepackage{amsfonts}       
\usepackage{nicefrac}       
\usepackage{microtype}      
\usepackage{xcolor}
\usepackage[noend]{algpseudocode}

\def\paperID{****}
\def\confName{CVPR}
\def\confYear{2026}

\title{Is Parameter Isolation Better for Prompt-Based Continual Learning?}

\author{
 Jiangyang Li$^{1}$\quad
 Chenhao Ding$^{1}$\quad 
 Songlin Dong$^{1,2}$\thanks{Corresponding author.}\quad
 Qiang Wang$^{1}$\quad 
 Jianchao Zhao$^{1}$\quad \\
 Yuhang He$^{1}$ \quad  
 Yihong Gong$^{1,2}$\quad
\\
$^1$Xi’an Jiaotong University \quad
$^2$Shenzhen University of Advanced Technology \quad
\\ 
}

\begin{document}
\maketitle
\begin{abstract}
Prompt-based continual learning methods effectively mitigate catastrophic forgetting. However, most existing methods assign a fixed set of prompts to each task, completely isolating knowledge across tasks and resulting in suboptimal parameter utilization. To address this, we consider the practical needs of continual learning and propose a prompt-sharing framework. This framework constructs a global prompt pool and introduces a task-aware gated routing mechanism that sparsely activates a subset of prompts to achieve dynamic decoupling and collaborative optimization of task-specific feature representations. Furthermore, we introduce a history-aware modulator that leverages cumulative prompt activation statistics to protect frequently used prompts from excessive updates, thereby mitigating inefficient parameter usage and knowledge forgetting. Extensive analysis and empirical results demonstrate that our approach consistently outperforms existing static allocation strategies in effectiveness and efficiency.
\end{abstract}    
\section{Introduction}
\label{sec:intro}

Continual learning (CL) \cite{aljundi2017expert, belouadah2021comprehensive} aims to endow models with the ability to learn new tasks sequentially without forgetting previously acquired knowledge. A persistent challenge in this setting is catastrophic forgetting \cite{mehta2023empirical, panos2023first, ramasesh2021effect, wang2024comprehensive, zhang2023slca}, where the model’s adaptation to new data overwrites past representations. Recently, prompt-based learning has emerged as a promising direction: instead of updating the entire network, it adapts to new tasks by optimizing a small number of task-specific prompts, while keeping the pre-trained backbone frozen \cite{li2021prefix, xin2024parameter, zhou2024continual}. This approach achieves parameter efficiency and naturally supports knowledge isolation, making it attractive for scalable lifelong learning.

Currently, mainstream prompt-based CL methods \cite{wang2022s, wang2022dualprompt, wang2023hierarchical, le2024mixture} are built on the assumption that tasks are independent and should be handled separately, and thus typically allocate a fixed set of prompts to each task. However, we rethink the practical demands of continual learning and argue that this paradigm has two main limitations. (i) Although this rigid allocation simplifies task management, it inherently limits parameter sharing between related tasks, struggles to handle the diverse capacity requirements of heterogeneous task sequences, and lacks the flexibility needed for scalable, long-term, continual learning. (ii) Static prompt assignments often result in inefficient use of model capacity, as different tasks may have varying degrees of relatedness or complexity that static prompts cannot easily accommodate.

\begin{figure*}[t]
\begin{center}
\centerline{\includegraphics[width=\textwidth]{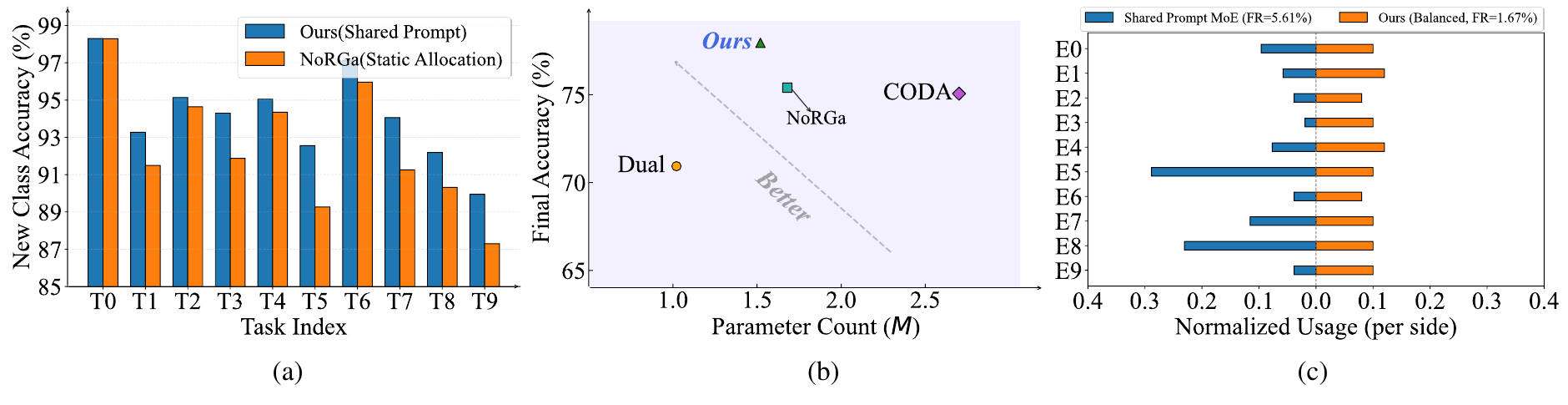}}
\caption{(a) Accuracy of novel classes at each incremental step on the CUB200 dataset. Our method achieves better novel class learning ability compared to static prompt allocation. (b) Comparison of parameter count and accuracy on the Imagenet-R dataset. Our method demonstrates a better accuracy–parameter trade-off than existing approaches. (c) Comparison of expert activation frequency distribution on the Imagenet-R dataset. Our proposed history-aware modulator achieves superior load balancing while effectively reducing knowledge forgetting~(FR).}
\label{fig:head}
\end{center}
\vspace{-10pt}
\end{figure*}

In this paper, we propose a \textbf{h}istory-\textbf{a}ware prompt-\textbf{sh}aring framework for continual learning (\textbf{Hash}). Specifically, we construct a global prompt pool, where each prompt functions as an expert specializing in certain feature patterns or task attributes. To enable efficient knowledge routing, we introduce a lightweight task-aware gating mechanism that dynamically selects a small subset of prompts for each input, facilitating adaptive composition of relevant knowledge. This parameter-sharing mechanism encourages positive transfer across related tasks and enables efficient local updates during continual learning. As shown in Figure~\ref{fig:head}(a), compared to static prompt allocation strategies, the proposed method demonstrates superior adaptability in learning new classes. Moreover, the mechanism supports flexible activation of prompt subsets based on task demands, providing strong capacity adaptivity. This sparse activation strategy promotes efficient parameter utilization and minimizes redundant computation across tasks. As illustrated in Figure~\ref{fig:head}(b), the proposed method achieves a better trade-off between accuracy and parameter efficiency, demonstrating enhanced scalability and computational efficiency.

While this shared prompting mechanism improves generalization on new classes, it may also cause certain prompts to be excessively updated due to repeated reuse, leading to catastrophic forgetting over time~\cite{gao2024beyond}. To strike a balance between accurate learning of new tasks and preserving previously acquired knowledge, we propose a history-aware modulator based on cumulative activation statistics. It consists of two components: (i) a routing penalty mechanism that reduces the selection scores of overused prompts to encourage balanced activation; and (ii) a gradient scaling strategy that attenuates the update magnitude for frequently activated prompts during training. These two components optimize the activation strategy of sub-prompt sets without altering the core architecture, enhancing parameter efficiency and effectively mitigating potential forgetting. As shown in Figure~\ref{fig:head}(c), the proposed history-aware modulator achieves better load balancing, and ablation studies demonstrate its significant reduction in forgetting rate.

Our contributions can be summarized as follows:
(1) We propose a continual learning framework based on a shared prompt pool with dynamic, sparse prompt selection, enabling flexible knowledge composition and improved parameter efficiency.
(2) We introduce a history-aware modulator that adaptively adjusts prompt selection and protects important prompts based on cumulative prompt activation statistics.
(3) We provide analytical and empirical evidence showing that shared prompt learning outperforms static prompt allocation in continual learning settings.
(4) \textbf{We conduct extensive experiments across diverse benchmarks, with comprehensive ablations and visualizations, consistently achieving superior performance over existing methods.}
\section{Related Work}
\textbf{Continual Learning:} Continual learning aims to enable models to learn sequential tasks without suffering from catastrophic forgetting. Rehearsal-based methods mitigate forgetting by storing and replaying representative samples from previous tasks~\cite{hou2019learning, rebuffi2017icarl, wang2022memory}, but raise concerns regarding memory efficiency and data privacy. A widely studied alternative is non-exemplar class-incremental learning, where no data from prior tasks is retained and task identifiers are not available during inference. To address this challenging setting, prior works have explored regularization-based constraints that penalize deviations from old parameters or impose knowledge distillation losses to preserve learned representations~\cite{li2017learning, huang2024etag, yu2020semantic}, augmentation strategies that synthesize pseudo-features or generate class prototypes from limited statistics to approximate old class distributions~\cite{zhu2021class, zhu2021prototype, kim2024cross}, and model rectification methods that adjust classifier outputs through bias correction or score calibration to mitigate prediction bias toward new classes~\cite{wang2023non}. Recently, approaches leveraging pre-trained models~\cite{dosovitskiy2020image} have shown promising results for non-exemplar CIL, exploiting rich transferable representations to achieve strong generalization without maintaining old-task memory. Among them, many adopt prompt-based tuning and employ instance-level prompt selection to dynamically retrieve task-relevant knowledge, effectively disentangling task-specific representations while reusing shared backbone knowledge~\cite{wang2022learning, wang2022dualprompt, smith2023coda, wang2023hierarchical, le2024mixture, gao2024consistent, roy2024convolutional}.

\noindent\textbf{Prompt Tuning: }Prompt tuning is a parameter-efficient paradigm for adapting pre-trained models to downstream tasks. Existing approaches can be broadly categorized into two groups: task-specific prompt allocation and shared prompt pooling. 
The first group assigns dedicated prompts to each task. DualPrompt~\cite{wang2022dualprompt}, CPrompt~\cite{gao2024consistent}, and HiDe-Prompt~\cite{wang2023hierarchical} statically allocate task-specific prompts to encode task identity, often combining them with shared components or contrastive regularization to improve generalization. NoRGa~\cite{le2024mixture} extends this idea by introducing a non-linear residual gating mechanism, viewing prompts as implicit experts to enhance flexibility. 
The second group explores shared prompt pools with dynamic selection. L2P~\cite{wang2022learning}, CODA-Prompt~\cite{smith2023coda}, and SMoP~\cite{choi2023smop} maintain a global pool of prompts and select subsets based on input features using learned routing mechanisms. CPG~\cite{lu2025training} also adopts a shared MoE-based prompt pool, but enforces consistency via orthogonal projection constraints on both router and experts to prevent representational drift. OVOR~\cite{huang2024ovor} takes a minimalist approach by using only a \emph{single shared prompt} across all tasks, relying instead on virtual outlier regularization applied to the classifier head to tighten decision boundaries and reduce inter-task confusion. ConvPrompt~\cite{roy2024convolutional} similarly uses a shared prompt pool but incorporates external knowledge by retrieving class-level attributes from GPT-3 during training, introducing additional supervision beyond standard continual learning settings and thus excluded from direct comparison. 
Building upon HiDe-Prompt~\cite{wang2023hierarchical}, which achieves high accuracy in task identification through hierarchical key-prompt matching, we introduce MoE into prompt tuning for enhanced specialization capabilities. However, this raises a critical challenge in continual learning: balancing forgetting and generalization. We propose history-aware expert modulation to address this, dynamically regulating expert contributions via cumulative usage statistics to maintain both backward compatibility and forward adaptability.
\begin{figure*}[t]
 \centering
\includegraphics[width=\textwidth]{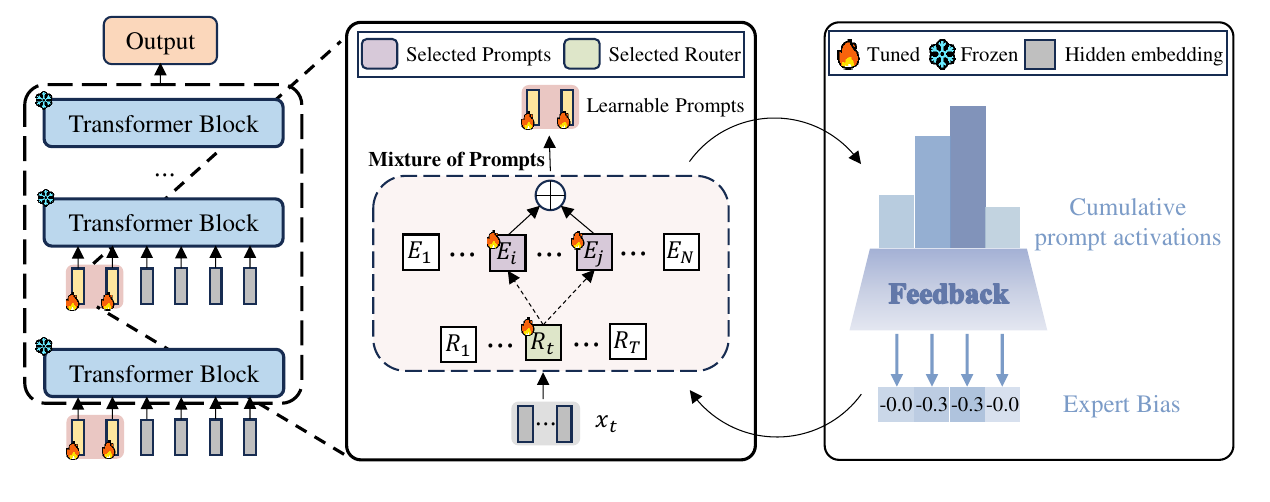}
\caption{Network architecture overview: Our method adopts a MoE-based shared prompt strategy. During training, the historical activation count of each prompt is recorded and incorporated as a bias term into the router’s scoring mechanism, enabling dynamic and informed expert selection based on historical usage.}
\label{fig:framework}
\end{figure*}
\section{Methods}\label{sec:method}
\subsection{Problem Definition}
We consider the task of \emph{non-exemplar class-incremental learning} (CIL), where a model must continually learn a sequence of classification tasks $\{\mathcal{T}_1, \mathcal{T}_2, \dots, \mathcal{T}_T\}$ without storing any data from previously encountered tasks. Each task $\mathcal{T}_t$ introduces a disjoint set of classes $\mathcal{C}_t$, and training data is only accessible during its corresponding learning phase. At test time, the model must jointly classify over the union of all seen classes $\mathcal{C}_{1:t} = \bigcup_{i=1}^t \mathcal{C}_i$ without access to the explicit task identifier. At stage $t$, the model receives a dataset $\mathcal{D}_t = \{(x_i^{(t)}, y_i^{(t)})\}_{i=1}^{N_t}$, where $x_i^{(t)}\in\mathbb{R}^{L\times d}$ of sequence length $L$ and embedding dimension $d$ denotes the input and $y_i^{(t)} \in \mathcal{C}_t$ the corresponding label. The model is optimized to minimize a supervised loss $\mathcal{L}_{\text{cls}}$ over $\mathcal{D}_t$ while maintaining performance on all prior tasks $\mathcal{T}_{1:t-1}$.

\subsection{Shared Prompt Pool with Dynamic Allocation}
To promote knowledge sharing across tasks, we introduce the shared prompt pool architecture based on Mixture-of-Experts, which dynamically selects prompts for each input to enhance generalization and reuse.

The prompt pool $\mathcal{P} = \{p_1, p_2, \dots, p_K\}$ contains $K$ prompts, each $p_k \in \mathbb{R}^{L_p \times d}$ of sequence length $L_p$ acting as an expert. To enable task-specific selection, we assign each of the $T$ tasks a dedicated router for dynamically selecting relevant prompts.

For a sample \( x_i^{(t)} \) from task \( \mathcal{T}_t \), its corresponding router \( \mathcal{R}_t \) computes the relevance score between the sample and each prompt \( p_k \) in the prompt pool using a weight matrix \( W_r \in \mathbb{R}^{d \times K} \) as follows:
\begin{align}
    \tilde{s}_{}^{(t)} = \mathcal{R}_t(x_i^{(t)}) = \frac{x_i^{(t)} W_r}{\sqrt{d}},
\end{align}
where $\sqrt{d}$ is a normalization factor. The score matrix $\tilde{s} \in \mathbb{R}^{L \times K}$ is averaged over the sequence length dimension $L$ to obtain a one-dimensional score vector $s = [s_1, s_2, \dots, s_K]$ of length $K$. The top-$k$ prompts are then selected based on $s$, denoted as $\mathcal{K}(s) = \text{TopK}(s)$. The normalized weights $\alpha_k$ for the selected prompts are computed as:
\begin{align}\label{eq:norm-weight}
    \alpha_k &= \frac{\exp(s_k)}{\sum_{j \in \mathcal{K}(s)} \exp(s_j)}, \quad k \in \mathcal{K}(s).
\end{align}
The instance-specific prompt $\tilde{p}$ is then computed as the weighted sum of the selected top-$k$ prompts:
\begin{align}\label{eq:weighted}
    \tilde{p} &= \sum_{k \in \mathcal{K}(s)} \alpha_k p_k.
\end{align}
To integrate the task-specific prompt $\tilde{p}$ into the attention mechanism, we divide it into two parts: $\tilde{p}^K, \tilde{p}^V \in \mathbb{R}^{\frac{L_p}{2} \times d}$, which are added to the key and value matrices, respectively. Although we omit the layer index for notational simplicity, a distinct instance-specific prompt $\tilde{p}^{(l)}$ is generated and injected into each attention layer $l$. Here, the splitting process is denoted as $\tilde{p}^K, \tilde{p}^V = \text{split}(\tilde{p})$.

Subsequently, we prepend $\tilde{p}^K$ and $\tilde{p}^V$ to the key and value matrices in the attention computation, respectively, resulting in the augmented matrices: $\mathbf{K}'^{(l)} = [\tilde{p}^K; \mathbf{K}^{(l)}]$ and $\mathbf{V}'^{(l)} = [\tilde{p}^V; \mathbf{V}^{(l)}]$. Accordingly, the final attention is computed as follows:
\begin{align}
    \text{Attn}(\mathbf{Q}^{(l)}, \mathbf{K}'^{(l)}, \mathbf{V}'^{(l)}) &= \text{softmax} \left( \frac{\mathbf{Q}^{(l)} (\mathbf{K}'^{(l)})^\top}{\sqrt{d_H}} \right) \mathbf{V}'^{(l)},
\end{align}
where $\mathbf{Q}^{(l)}$, $\mathbf{K}^{(l)}$, and $\mathbf{V}^{(l)}$ represent the query, key, and value matrices for layer $l$, respectively, and $d_H$ is the dimensionality of the attention heads. By integrating dynamically composed prompts into the attention backbone, our framework unifies specialization and generalization in a single, efficient mechanism.

Benefiting from our shared prompt pool and dynamic allocation mechanism, our method achieves a lower total parameter count than Hide-Prompt, even when including our router. See Appendix for detailed analysis.

\noindent\textbf{Analysis: Benefits of Prompt Sharing.} Unlike static methods that allocate disjoint prompt subsets to each task, our approach constructs a continuous and expressive representation space by dynamically composing prompts from a global pool. Formally, for static allocation where each task $\mathcal{T}_t$ uses a fixed subset $\mathcal{P}_t = \{p_{t,1}, \dots, p_{t,k}\} \subset \mathcal{P}$, the optimization objective is:
\begin{align}
\mathcal{L}^{\text{static}}_t = \mathcal{L}_{\text{cls}}(f(x; \theta, \mathcal{P}_t)),
\end{align}
yielding isolated updates with no inter-task parameter sharing. In contrast, our method leverages a shared pool $\mathcal{P} = \{p_1, \dots, p_K\}$ and constructs input-adaptive prompts $\tilde{p}(x)$ via relevance-weighted composition:
\begin{align}
\tilde{p}(x) = \sum_{k \in \mathcal{K}(s)} \alpha_k p_k,\quad 
\mathcal{L}^{\text{moe}}_t = \mathcal{L}_{\text{cls}}(f(x; \theta, \tilde{p}(x))),
\end{align}

This design induces a smooth and compositional prompt space that facilitates prompt reuse, expands representational capacity, and enhances generalization to novel classes. 

The entropy $H(\alpha) = -\sum_k \alpha_k \log \alpha_k$ quantifies the diversity of prompt activation for each input. A higher $H(\alpha)$ implies that the model distributes attention across multiple prompts to accommodate input complexity, while a lower $H(\alpha)$ reflects selective activation for simpler prompts. This input-aware modulation supports a flexible allocation of capacity, allowing the model to generalize across tasks while retaining sufficient specialization where needed.

\subsection{History-Aware Modulator}
\label{sec:dynamic_routing}
To further address the issue of prompt overuse in traditional shared prompt pool methods~\cite{gao2024beyond}, we introduce a history-aware mechanism which dynamically adjusts expert routing scores and gradient update magnitudes based on historical usage statistics, thereby mitigating the overuse of prompt and alleviating catastrophic forgetting.

\begin{table*}
  \centering
  \caption{Overall performance comparison on Split CIFAR-100 and Split ImageNet-R. We present Final Average Accuracy (FAA), Cumulative Average Accuracy (CAA), and Forgetting Measure (FM) of all methods. ``Shared'' indicates whether the method adopts a shared prompt strategy.}
\small
\setlength{\tabcolsep}{7pt}
\begin{tabular}{lcllllll}
\toprule
\multicolumn{1}{c}{\multirow{2}{*}{Method}} & \multicolumn{1}{c}{\multirow{2}{*}{Shared}} & \multicolumn{3}{c}{Split CIFAR-100} & \multicolumn{3}{c}{Split Imagenet-R} \\ 
\cmidrule(l){3-5} \cmidrule(r){6-8}
\multicolumn{1}{c}{} & \multicolumn{1}{c}{} & \multicolumn{1}{c}{\textbf{FAA} ($\uparrow$)} & \multicolumn{1}{c}{\textbf{CAA}($\uparrow$)} & \multicolumn{1}{c}{FM($\downarrow$)} & \multicolumn{1}{c}{\textbf{FAA} ($\uparrow$)} & \multicolumn{1}{c}{\textbf{CAA}($\uparrow$)} & \multicolumn{1}{c}{FM($\downarrow$)} \\ 
\midrule
DualPrompt\cite{wang2022dualprompt} & $\times$ & $87.30 \pm 0.27$ & $91.23 \pm 0.65$ & $3.87 \pm 0.43$ & $70.93 \pm 0.08$ & $75.67 \pm 0.52$ & $5.47 \pm 0.19$ \\
CPrompt~\cite{gao2024consistent} & $\times$ & $87.82 \pm 0.21$ & $92.53 \pm 0.23$ & $5.06 \pm 0.50$ & $77.15 \pm 0.11$ & $82.92 \pm 0.70$ & $5.97 \pm 0.68$ \\
HiDe-Prompt~\cite{wang2023hierarchical} & $\times$ & $92.61 \pm 0.28$ & $94.03 \pm 0.01$ & $1.50 \pm 0.28$ & $75.06 \pm 0.12$ & $76.60 \pm 0.01$ & $4.09 \pm 0.13$ \\
NoRGa~\cite{le2024mixture} & $\times$ & $94.48 \pm 0.13$ & $95.83 \pm 0.37$ & $\textbf{1.44} \pm 0.27$ & $75.40 \pm 0.39$ & $79.52 \pm 0.07$ & $4.59 \pm 0.07$\\ 
EvoPrompt~\cite{kurniawan2024evolving} & $\times$ & $87.97 \pm 0.30$ & $89.12 \pm 0.21$ & $2.60 \pm 0.42$ & $76.83 \pm 0.08$ & $78.43 \pm 0.42$ & $2.78 \pm 0.06$ \\
\midrule
L2P~\cite{wang2022learning} & $\checkmark$ & $83.06 \pm 0.17$ & $88.27 \pm 0.71$ & $5.61 \pm 0.32$ & $67.53 \pm 0.44$ & $71.98 \pm 0.52$ & $5.84 \pm 0.38$ \\
CODA-Prompt~\cite{smith2023coda} & $\checkmark$ & $86.94 \pm 0.63$ & $91.57 \pm 0.75$ & $4.04 \pm 0.18$ & $70.03 \pm 0.47$ & $74.26 \pm 0.24$ & $5.17 \pm 0.22$ \\
OVOR~\cite{huang2024ovor} & $\checkmark$ & $85.99 \pm 0.21$ & $90.26 \pm 0.35$ & $6.42 \pm 0.17$ & $76.11 \pm 0.33$ & $79.12 \pm 0.28$ & $7.16 \pm 0.28$ \\
CPG~\cite{lu2025training} & $\checkmark$ &$90.63 \pm 0.44$ & $93.24 \pm 0.33$ & $3.98 \pm 0.65$ & $78.63 \pm 0.52$& $81.04 \pm 0.23$ & $7.18 \pm 0.62$ \\

Hash~(Ours) & $\checkmark$ & $\textbf{95.02} \pm 0.34$ & $\textbf{95.97} \pm 0.28$ & $1.67 \pm 0.23$ & $\textbf{79.02} \pm 0.36$ & $\textbf{82.96} \pm 0.17$ & $\textbf{2.63} \pm 0.15$\\ 

\bottomrule
\end{tabular}
\label{tab:cifar_imagenet}
\end{table*}

\noindent\textbf{History-Aware Dynamic Routing.} In the \textbf{h}istory-aware \textbf{d}ynamic \textbf{r}outing~(HDR) mechanism, we record the cumulative activation count $\mathcal{H}_e^t$ for the $e$-th prompt up to task $t$, which is then used for load balancing within the prompt pool. Specifically, for any input, each prompt is assigned an initial relevance score $s_e$. To prevent the overuse of certain prompts, we apply a penalization function $\psi: \mathbb{R}_+ \to \mathbb{R}_+$ to their activation history:
\begin{align}
    \tilde{s}_e = s_e - \psi(\mathcal{H}_e^t),
\end{align}
where $\tilde{s}_e$ is the updated score. The penalty function $\psi$ may be logarithmic (e.g., $\log(1 + \mathcal{H}_e^t)$), polynomial (e.g., $(\mathcal{H}_e^t)^\beta
$), or even task-specific. Ablation studies for this setting are provided in the appendix. In our default implementation, we adopt a simple stepwise penalization applied to top-$k$ activated experts:
\begin{align}
    \tilde{s}_e =
    \begin{cases}
        s_e - \delta, & \text{if } e \in \mathcal{A}_t, \\
        s_e, & \text{otherwise},
    \end{cases}
\end{align}
where $\delta > 0$ is a fixed deduction and $\mathcal{A}_t$ denotes the top-$k$ experts with the highest cumulative activation count $\mathcal{H}_e^t$ up to task $t$. This can be seen as a discrete approximation of a continuous penalization function, enforcing sparse regularization in a greedy fashion.

\noindent\textbf{History-Aware Gradient Modulation.} To further preserve the historical knowledge of prompts and prevent its disruption in subsequent learning, we propose \textbf{h}istory-aware \textbf{g}radient \textbf{m}odulation~(HGM) to protect prompts that have been activated in the past. Specifically, we use a monotonic decay function $\gamma: \mathbb{R}_+ \to (0, 1]$ to modulate the gradients. 
\begin{align}
    \tilde{g}_e = \gamma(\mathcal{H}_e^t) \cdot g_e,
\end{align}
where $g_e$ is the raw gradient for expert $e$. The modulation function can take various forms, such as inverse scaling $\gamma(h) = 1 / (1 + \beta h)$ or exponential decay $\exp(-\beta h)$, with $\beta > 0$. In practice, we use a piecewise constant modulation for the top-$k$ experts:
\begin{align}
    \gamma(\mathcal{H}_e^t) =
    \begin{cases}
        \alpha, & \text{if } e \in \mathcal{A}_t, \\
        1, & \text{otherwise},
    \end{cases}
\end{align}
with $0 < \alpha < 1$. This reduces the learning rate for frequently activated experts, implicitly enforcing parameter stability. Ablation studies on different forms of $\gamma(h)$ are provided in the appendix.

\noindent\textbf{History-Aware Modulator from a Regularization Perspective.} We aim to reconsider the problem from a regularization perspective, formalizing the trade-off between maximizing expert scores and penalizing their overuse. This unified framework not only explains our dynamic expert routing and history-aware gradient modulation but also validates our design intuition.

The overall mechanism can be interpreted through the lens of regularized optimization. Expert routing aims to maximize the score while penalizing overused experts:
\begin{align}
    \min_{\mathbf{p} \in \Delta^E} \ \mathcal{L}_{\text{route}}(\mathbf{p}) = \sum_{e=1}^E p_e \cdot (-s_e) + \sum_{e=1}^E p_e \cdot \psi(\mathcal{H}_e^t),
\end{align}
where $\Delta^E$ is the probability simplex. While the full minimization corresponds to soft attention over all experts, our top-$k$ routing can be viewed as a sparse approximation under entropic regularization.

For gradient modulation, our update step approximates the minimization of a stability-aware regularized loss:
\begin{align}
    \min_{\theta} \ \mathcal{L}(\theta) + \sum_{e=1}^E \underbrace{R_e(\theta_e; \mathcal{H}_e^t)}_{\text{history-aware regularization}},
\end{align}
where $R_e$ penalizes deviation from prior parameters $\theta_e^t$ according to historical activation:
\begin{align}
    R_e(\theta_e; \mathcal{H}_e^t) = \frac{1}{2} \cdot \mathcal{H}_e^t \cdot \|\theta_e - \theta_e^t\|^2.
\end{align}
This induces a stability constraint that grows with expert usage, aligning with the intuition of selectively freezing experts that are heavily relied upon.

By embedding historical usage into both expert routing and gradient modulation, our mechanism strikes a dynamic balance between adaptability and stability. This approach enables continuous control over routing selection and gradient flow, while inherently maintaining robustness in response to diverse inputs. 

\section{Experiments}\label{sec:exp}
\subsection{Experemental Details}
\noindent\textbf{Benchmark.} We evaluate our method on a suite of diverse class-incremental learning (CIL) benchmarks to assess its adaptability across various scenarios. \textit{Split CIFAR-100} comprises 100 classes of natural images~\cite{krizhevsky2009learning}, partitioned into 10 sequential tasks with disjoint class labels. \textit{Split ImageNet-R} is constructed from 200 classes sampled from ImageNet-R~\cite{hendrycks2021many}, including renditions, sketches, and abstract styles. These are divided into 10 tasks and present a significant domain shift. \textit{Split CUB-200} is derived from the CUB-200-2011 dataset~\cite{wah2011caltech}, a fine-grained bird classification benchmark split into 10 tasks. 5-Datasets \cite{ebrahimi2020adversarial} comprises CIFAR-10 \cite{krizhevsky2009learning}, MNIST \cite{lecun1998gradient}, Fashion-MNIST \cite{xiao2017fashion}, SVHN \cite{netzer2011reading}, and notMNIST \cite{bulatov2011notmnist}, each serving as a distinct incremental task to evaluate the model's robustness under significant inter-task distributional shifts.

\begin{table}[ht]
  \centering
  \caption{The results on Split CUB-200 and 5-Datasets. Here we present FAA and CAA for all approaches.}\label{tab:cub}
  \small
  \setlength{\tabcolsep}{10pt}
    \begin{tabular}{lcccc}
    \toprule
    \multirow{2}{*}{Method} & \multicolumn{2}{c}{Split CUB-200} & \multicolumn{2}{c}{5-Datasets} \\
    \cmidrule(l){2-3} \cmidrule(r){4-5}
     & FAA & CAA & FAA & CAA \\ \midrule
    L2P         & 75.46 & 81.59 & 81.84 & 85.36 \\
    DualPrompt  & 77.56 & 83.24 & 77.91 & 84.92 \\
    CODA-Prompt & 74.34 & 79.52 & 64.18 & 77.62  \\
    CPrompt     & 80.35 & 87.66 & 84.96 & 88.67 \\
    HiDe-Prompt & 86.56 & 89.34 & 93.83 & 94.32 \\
    NoRGa       & 90.90 & 92.53 & 94.16 & 94.55 \\
    Hash~(Ours)        & \textbf{91.34} & \textbf{92.98} & \textbf{95.12} & \textbf{95.67} \\
    \bottomrule
    \end{tabular}
\end{table}
\begin{table}
  \centering
  \caption{Long sequence incremental analysis: Experimental results with 20 incremental tasks.}
\small
\setlength{\tabcolsep}{8pt}
\begin{tabular}{lcccc}
\toprule
\multicolumn{1}{c}{\multirow{2}{*}{Method}} & \multicolumn{2}{c}{Split CIFAR-100} & \multicolumn{2}{c}{Split ImageNet-R} \\ \cmidrule(l){2-3} \cmidrule(r){4-5}
\multicolumn{1}{c}{} & \multicolumn{1}{c}{FAA} & \multicolumn{1}{c}{FM} & \multicolumn{1}{c}{FAA} & \multicolumn{1}{c}{FM} \\ \midrule
L2P         & 79.95 & 5.69 & 69.72 & 5.92 \\
DualPrompt  & 81.34 & 4.02 & 66.72 & 5.63 \\
CODA-Prompt & 81.94 & 4.92 & 69.96 & 6.79 \\
CPrompt & 84.57 & 6.01 & 74.79 & 7.34 \\
HiDe-Prompt & 91.68 & 2.34 & 73.98 & 4.98 \\
NoRGa & 93.12 & 2.12 & 74.32 & 5.14 \\

\textbf{Hash~(Ours)} & \textbf{94.07} & \textbf{1.98} & \textbf{76.57} & \textbf{4.62}  \\
\bottomrule
\end{tabular}
\label{tab:long}
\end{table}

\noindent\textbf{Baselines.} We compare our method with a suite of representative prompt-based continual learning approaches, including L2P \cite{wang2022learning}, DualPrompt \cite{wang2022dualprompt}, CODA-Prompt \cite{smith2023coda}, C-Prompt \cite{gao2024consistent}, HiDe-Prompt \cite{wang2023hierarchical}, and NoRGa \cite{le2024mixture}, covering diverse designs for prompt integration and selection. All methods are evaluated under consistent experimental settings using the same backbone and training pipeline. The main paper reports results under the supervised ImageNet-21K (Sup-21K) \cite{ridnik2021imagenet} pre-training, while \textbf{results under additional pre-training paradigms (iBOT \cite{zhouimage}, DINO \cite{caron2021emerging}, MoCo \cite{chen2021empirical}) are deferred to the Appendix for completeness.}

\noindent\textbf{Evaluation Metrics.} Following standard practice, we report three widely-used metrics to evaluate continual learning performance: final average accuracy (FAA), cumulative average accuracy (CAA), and final forgetting measure (FM) \cite{wang2024comprehensive}. FAA reflects the average performance on all tasks after learning the final task, while CAA summarizes the historical performance across all seen tasks. FM quantifies the extent of knowledge forgetting between tasks. \textbf{The detailed computation of evaluation metrics is provided in the Appendix.}

\noindent\textbf{Implementation Details.} All experiments are conducted on a single NVIDIA A100 GPU. We adopt the pre-trained ViT-B/16 as the backbone across all methods. Training is performed using the Adam optimizer with default parameters ($\beta_1 = 0.9$, $\beta_2 = 0.999$), a batch size of 128. We adopt the same epoch configuration for all methods as provided by HiDe-Prompt~\cite{wang2023hierarchical}. Prompt selection is performed at the \textbf{instance level} during both training and inference. We empirically verify that the test results are \textbf{invariant} to batch size (Appendix). All results are averaged over three random seeds.
\textbf{Additional details on hyperparameter configurations can be found in the Appendix.}

\subsection{Comparison Results}
\noindent\textbf{Class-IL Results.} Table~\ref{tab:cifar_imagenet} reports results on Split CIFAR-100 and Split ImageNet-R under the Sup-21K pre-trained vision transformer. Our method achieves the highest FAA and CAA across both benchmarks, while maintaining competitive FM, demonstrating strong performance and stability in class-incremental learning. Specifically, we obtain 95.02\% FAA on CIFAR-100 and 79.02\% on ImageNet-R, outperforming NoRGa by 0.54\% and 3.66\%, respectively. Notably, while regularization-based methods typically achieve low FM, our design strikes a better balance between final accuracy and forgetting, validating the effectiveness of expert routing and gradient modulation. Table~\ref{tab:cub} further shows performance on Split CUB-200 and 5-Datasets benchmark. Again, our method achieves the best FAA of 91.34\% on CUB-200 and 95.12\% on 5-Datasets, outperforming the strongest baseline by 0.44\% and 0.96\%, respectively. These improvements are particularly meaningful in settings prone to overfitting or under-generalization. Overall, the consistent performance gains across different datasets and pre-training types validate the effectiveness and adaptability of our MoE-enhanced dynamic prompt learning framework.
\textbf{More results are presented in the Appendix, including comparisons with other recent adapter-based approaches.}

\begin{table}[t]
	\centering
    \caption{Results on CORe50. $A_N$ gives the accuracy averaged over tasks and buffer size represents the number of old exemplars stored in the memory buffer.}
    \small
    \setlength{\tabcolsep}{10pt}
	\begin{tabular}{c|c|c}
            \hline
            \rule{0pt}{9pt} Method  & Buffer size & $A_N$ ($\uparrow$) \\
            \hline
            ER   & & $80.10$ \\ 
            GDumb        &  & $74.92$   \\  
            BiC      &  & $79.28$   \\ 
            DER++    & \textbf{50/class} & $79.70$   \\ 
            Co$^2$L     &  & $79.75$   \\
            DyTox &  & $79.21$ \\  
            L2P &  & $81.07$  \\
            \hline
            EWC      & & $74.82$   \\ 
            LwF    & & $75.45$   \\ 
            L2P     &\textbf{0/class} & $78.33$   \\
            CODA-P-S &  & $85.41$ \\  
            S-Prompts &  & $83.13$  \\
            PINA &  & 86.74 \\
            \textbf{Hash~(Ours)}&  & \textbf{87.94}  \\
            \hline
        \end{tabular}
        \label{tab:core50}
\end{table}
\begin{table}[t]
	\centering
    \caption{Ablation study on key components of the proposed method.} 
    \small
  \setlength{\tabcolsep}{4.7pt}
	\begin{tabular}{ccccccc}
	 \toprule
      \multicolumn{1}{c}{\multirow{2}{*}{MoE}} & \multicolumn{1}{c}{\multirow{2}{*}{HDR}} & \multicolumn{1}{c}{\multirow{2}{*}{HGM}} & \multicolumn{2}{c}{Split CIFAR-100} & \multicolumn{2}{c}{Split Imagenet-R} \\ \cmidrule(l){4-5} \cmidrule(r){6-7}
      \multicolumn{1}{c}{} & \multicolumn{1}{c}{} & \multicolumn{1}{c}{} & \multicolumn{1}{c}{FAA} & \multicolumn{1}{c}{CAA} & \multicolumn{1}{c}{FAA} & \multicolumn{1}{c}{CAA} \\ \midrule
       \checkmark & & &88.56 &90.12 &76.99 &79.53  \\ 
       \checkmark &\checkmark & &94.15 &95.03 &78.10 &81.21 \\ 
       \checkmark & &\checkmark &93.86 &94.77 &77.56 &80.82 \\ 

       \checkmark &\checkmark &\checkmark &\textbf{95.02} &\textbf{95.97} &\textbf{79.02} &\textbf{82.96} \\ 
       \hline
	\end{tabular}
	\label{tab:ablation}
\end{table}
\begin{table}[t]
\centering
\caption{Performance comparison with different prompt positions.}
\small
\setlength{\tabcolsep}{6pt}
\begin{tabular}{ll|cccc}
\toprule
\multicolumn{2}{c}{\multirow{2}{*}{Hash}} & \multicolumn{2}{c}{Split CIFAR-100} & \multicolumn{2}{c}{Split ImageNet-R} \\
\cmidrule{3-6}
\multicolumn{2}{c}{} & FAA & CAA & FAA & CAA \\
\midrule
(Default) & 1--4     & \textbf{95.02} & \textbf{95.97} & \textbf{79.02} & \textbf{82.96} \\
\midrule
\multirow{2}{*}{Deep} 
    & 5--12        & 94.67 & 95.59 & 78.61 & 82.48  \\
    & 1--12        & 94.81 & 95.76 & 78.75 & 82.65  \\
\midrule
\multirow{2}{*}{Shallow} 
    & 1           & 93.28 & 94.12 & 77.00 & 80.94  \\
    & 12          & 93.01 & 93.85 & 76.82 & 80.70  \\
\bottomrule
\end{tabular}
\label{tab:layer}
\end{table}

\noindent\textbf{Long Sequence Incremental Analysis.} We assess the performance of all methods under a more challenging long-sequence incremental learning setup, where each dataset is uniformly divided into 20 sessions. All methods are evaluated using the ViT-B/16 backbone pre-trained on ImageNet-21K. Compared to short-session setups, this setting imposes greater pressure on knowledge retention and prompt scalability due to finer task granularity, increased semantic overlap, and extended task sessions. These challenges tend to amplify catastrophic forgetting in existing methods. As shown in Table~\ref{tab:long}, our method consistently maintains high accuracy and low forgetting across both CIFAR-100 and ImageNet-R. Notably, the performance gap widens with increasing task length, indicating that our approach scales more effectively under prolonged incremental learning. This further highlights the advantage of our shared, input-adaptive prompt mechanism, which promotes efficient knowledge reuse and mitigates catastrophic forgetting even in the presence of long and complex task sequences.
\textbf{Additional results under an even more extreme setting with 50 incremental tasks are provided in Appendix.}

\noindent\textbf{Domain-IL Results.} To complement our evaluation under class-incremental settings, we further assess the performance of our approach in the domain-incremental learning (DIL) scenario, where tasks share the same label space but differ in domain-specific input distributions. We compare our method with three recent DIL baselines: CODA~\cite{smith2023coda}, S-Prompts~\cite{wang2022s}, and PINA~\cite{wang2024non}. CODA employs domain-specific adapters with contrastive objectives to separate domain-invariant and domain-specific features, while S-Prompts maintains a per-domain prompt pool with a learned scoring mechanism for selection. PINA, in contrast, introduces a cross-domain concept integration framework that effectively aggregates knowledge from multiple domains, demonstrating strong performance on several DIL benchmarks. We report results on CORe50 in Table~\ref{tab:core50}, showing that our method outperforms CODA by +2.53\% and surpasses S-Prompt, which was specifically tailored for DIL. Notably, our approach also exceeds the performance of PINA by a margin of 1.2\%, further highlighting its effectiveness. All methods are evaluated using the same pre-trained backbone (ViT-B/16), and exemplar-based approaches store 50 exemplars per class following standard protocol. These findings highlight the robustness and generality of our approach across diverse continual learning settings.

\begin{figure*}[t!]
\begin{center}
\centerline{\includegraphics[width=\textwidth]{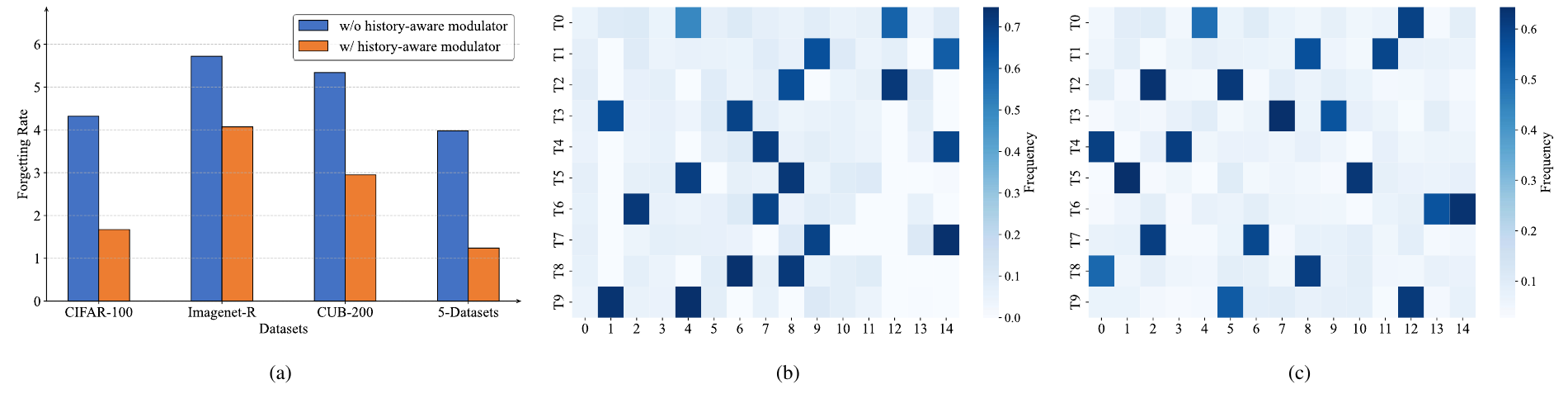}}
\caption{(a) Forgetting rate reduction with history-aware modulator. (b) Cross-task expert utilization frequency distribution without history-aware modulator. (c) Cross-task expert utilization frequency distribution with history-aware modulator.}
\label{fig:balance}
\end{center}
\vspace{-10pt}
\end{figure*}

\subsection{Ablation Study}
\noindent\textbf{Impact of History-Aware Modulator.} To evaluate the effectiveness of our proposed history-aware routing mechanism, we conduct a comparative analysis of forgetting rates across multiple datasets. Figure~\ref{fig:balance} illustrates the impact of our approach on catastrophic forgetting mitigation. Figure~\ref{fig:balance}(a) shows that history-aware routing consistently reduces forgetting rates across all datasets, with absolute decreases of 2.65\%, 1.65\%, 2.39\%, and 2.74\% on CIFAR-100, Imagenet-R, CUB-200, and 5-Datasets, respectively. This stems from two complementary synergistic effects: routing penalties prevent over-reliance on dominant prompts, while gradient scaling protects their learned representations from excessive updates. Figure~\ref{fig:balance}(b) shows that without the modulator, expert usage is highly skewed, with a strong bias toward a few dominant experts. After applying the history-aware modulator (Figure~\ref{fig:balance}(c)), expert utilization becomes significantly more balanced without sacrificing cross-task sharing, thus improving the overall efficiency of the shared prompt pool. The activation pattern shifts from highly concentrated to broadly distributed, ensuring more diverse prompt combinations across tasks.

\begin{table}[t]
  \centering
  \caption{Sensitivity of the number of router experts.}
  \small
  \setlength{\tabcolsep}{14pt}
    \begin{tabular}{ccccc}
	 \toprule
      \multicolumn{1}{c}{\multirow{2}{*}{$N_r$}} & \multicolumn{2}{c}{Split CIFAR-100} & \multicolumn{2}{c}{Split Imagenet-R} \\ \cmidrule(l){2-3} \cmidrule(r){4-5}
      \multicolumn{1}{c}{} & \multicolumn{1}{c}{FAA} & \multicolumn{1}{c}{CAA} & \multicolumn{1}{c}{FAA} & \multicolumn{1}{c}{CAA} \\ \midrule
       10 &94.89 &95.32 &78.76 &82.54 \\ 
       15 &95.02 &95.97 &79.02 &82.96 \\ 
       20 &95.08 &96.14 &78.95 &82.72 \\ 
       25 &94.54 &95.08 &78.28 &81.92  \\ 
       \hline
	\end{tabular}
    \label{tab:N_r}
\end{table}

\noindent\textbf{Module-Wise Ablation.} To disentangle the contributions of our two key components, namely history-aware dynamic routing (HDR) and history-aware gradient modulation (HGM), we perform an ablation study summarized in Table~\ref{tab:ablation}. The study reports the FAA and CAA. Beginning with a naive MoE-based prompt tuning baseline with uniform expert assignment, we first introduce HDR alone, which improves accuracy from 88.56\% to 94.15\% on Split CIFAR-100 and from 76.99\% to 78.10\% on Split Imagenet-R under the Sup-21K backbone, yielding relative gains of 5.59\% and 1.11\% respectively. This confirms that leveraging historical expert activations enhances expert selection efficacy. Applying HGM in isolation achieves similar improvements—an increase of 0.57\% on Imagenet-R (Sup-21K), reaching 77.56\%, and a gain of 5.30\% on CIFAR-100. Finally, combining HDR and HGM yields the best results across all configurations, achieving 95.02\% on CIFAR-100 and 79.02\% on Imagenet-R, which corresponds to total gains of 6.46\% and 2.03\% over the naive MoE variant. These improvements underscore the complementary nature of the two components: HDR optimizes \emph{which} experts to activate, while HGM regulates \emph{how} they are updated, jointly enhancing both knowledge acquisition and retention.

\noindent\textbf{Layer-wise Prompt Deployment Analysis.} To investigate the impact of layer-wise prompt deployment, we systematically placed the prompt modules at different layers of the backbone and evaluated their effect. As shown in Table~\ref{tab:layer}, deploying prompts at the lower layers consistently yields the best results across benchmarks, achieving 95.02\% FAA on CIFAR-100 and 79.02\% on ImageNet-R. Notably, using deeper layers or all layers degrades performance by 0.21--0.41\%, while single-layer deployment performs significantly worse with drops of 2.02--2.20\%. This indicates that prompt-based adaptation is most effective when applied to early feature representations, where multi-layer coordination enables better balance between plasticity and stability. Lower layers encode more task-agnostic patterns that are easier to modulate without disrupting prior knowledge, whereas deeper features are more task-specific and less adaptable to continual learning scenarios.

\begin{table}[t]
  \centering
  \caption{Sensitivity of prompt lengths.}
  \small
  \setlength{\tabcolsep}{14pt}
	\begin{tabular}{ccccc}
	 \toprule
      \multicolumn{1}{c}{\multirow{2}{*}{$L$}} & \multicolumn{2}{c}{Split CIFAR-100} & \multicolumn{2}{c}{Split Imagenet-R} \\ \cmidrule(l){2-3} \cmidrule(r){4-5}
      \multicolumn{1}{c}{} & \multicolumn{1}{c}{FAA} & \multicolumn{1}{c}{CAA} & \multicolumn{1}{c}{FAA} & \multicolumn{1}{c}{CAA} \\ \midrule
       5 &94.12 &94.72 &77.44 &81.22 \\ 
       10 &94.95 &95.72 &78.55 &82.44 \\ 
       15 &95.02 &95.97 &79.02 &82.96 \\ 
       20 &95.11 &96.03 &78.39 &82.21 \\ 
       25 &94.79 &95.88 &78.32 &82.04 \\ 
       \hline
	\end{tabular}
    \label{tab:L}
\end{table}

\noindent\textbf{Hyperparameter Sensitivity Analysis.} We analyze the sensitivity of our method to the number of experts $N_r$ and the prompt length $L$, as reported in Table~\ref{tab:N_r} and Table~\ref{tab:L}. The performance remains relatively stable across different configurations, indicating the robustness of the method. For the number of experts, increasing $N_r$ generally leads to performance improvement up to a certain point, with optimal results achieved at $N_r=15-20$. When $N_r$ becomes too large (e.g., 25), the overall accuracy slightly declines, possibly due to insufficient training samples per expert or increased optimization difficulty. This highlights the importance of balancing model capacity and expert coverage. In terms of prompt length, longer prompts improve performance initially, with $L=15$ emerging as the optimal configuration across both benchmarks. Beyond this point, the benefit quickly saturates and may even slightly decline, suggesting that moderate lengths suffice for adaptation without incurring unnecessary memory cost.
\textbf{More results about the history-aware modulator are provided in the Appendix.}
\section{Conclusion}
This study presents a prompt-sharing framework for continual learning, which adopts a sparse and dynamic prompt allocation strategy. By enabling prompt sharing, our approach inherently promotes generalization and reduces redundancy compared to static prompt allocation designs. To address catastrophic forgetting, we further introduce a history-aware modulator which jointly balance knowledge transfer and task stability. Empirically, we show that it strikes an effective balance between generalization and forgetting, outperforming parameter-isolation prompt-based approaches.

\noindent\textbf{Limitations.} The current study is limited by computational resources, and we leave the exploration of larger-scale backbone models to future work.

{
    \small
    \bibliographystyle{ieeenat_fullname}
    \bibliography{main}
}
\clearpage
\setcounter{page}{1}
\maketitlesupplementary
\begin{algorithm*}[tb]
    \caption{Training Algorithm of Hash}
    \label{alg:hash}
    \textbf{Input}: Pre-trained transformer backbone $f_{\theta}$, training sets $\mathcal{D}_1, ..., \mathcal{D}_T$, number of tasks $T$, number of epochs $E$, hyperparameters $\tau$ and $\lambda$. \\ 
    \textbf{Output}: Parameters ${R}_{1}, ..., {R}_{T}$, ${p}_{1}, ..., {p}_{K}$, $\omega$ and $\psi$
    \begin{algorithmic}[1] 
          \State Initialize $R_{1}$, ${p}_{1}, ..., {p}_{K}$, $\omega$ and $\psi$
          \For{$t = 1, ..., T$}
              \For{$c \in \mathcal{Y}_t$} 
              \State Obtain $\hat{\mathcal{G}}_c$ from $f_{\theta}$ and $\mathcal{D}_t$ \Comment{Uninstructed Representations}
              \EndFor
              \State Initialize empty prompt batch: $\tilde{P}_t$
              \For{$x_i \in \mathcal{D}_t$}
                \State Construct routing score: $s_i = \mathcal{R}_t(x_i)$
                \State Construct top-$k$ index set: $\mathcal{K}(s_i) = \text{TopK}(s_i)$
                \State Construct attention weights: $\alpha_k = \exp(s_{i,k}) \big/ \sum_{j \in \mathcal{K}(h_i)} \exp(s_{i,j})$
                \State Construct instance prompt: $\tilde{p}_i = \sum_{k \in \mathcal{K}(s_i)} \alpha_k p_k$
                \State Append $\tilde{p}_i$ to $\tilde{P}_t$
              \EndFor
              \For{$epoch = 1, ..., E$} 
              \State Optimize $\boldsymbol{p}_{t}$ and $\psi$ with $\mathcal{L}_{{\rm{WTP}}}$ in Eq.~(\ref{eq:wtp_loss}) \Comment{Within-Task Prediction}
              \State Optimize $\omega$ with $\mathcal{L}_{{\rm{TII}}}$ in Eq.~(\ref{eq:tii_loss}) \Comment{Task-Identity Inference}
               \State Optimize $\psi$ with $\mathcal{L}_{{\rm{TAP}}}$ in Eq.~(\ref{eq:tap_loss}) \Comment{Task-Adaptive Prediction}
              \EndFor

              \For{$c \in \mathcal{Y}_t$} 
              \State Obtain $\mathcal{G}_c$ from $f_{\theta}$, $\tilde{P}_t$ and $\mathcal{D}_t$ \Comment{Instructed Representations} 
              \EndFor
                   
        \EndFor
       \State \textbf{return} $({R}_{1}, ..., {R}_{T}$, ${p}_{1}, ..., {p}_{K}$, $\omega$, $\psi)$
    \end{algorithmic}
\end{algorithm*}

\section{Training Algorithm}
We adopt the HiDe-Prompt framework \cite{wang2023hierarchical} and adjust it to suit our shared prompt pool setting. During the Within-Task Prediction (WTP) phase, we jointly optimize the classification head $\psi$ and the global prompt pool $\mathcal{P} = \{p_1, \dots, p_K\}$, where each training instance dynamically activates a sparse subset of prompts via a learned task-specific router. The selected prompt composition $\tilde{p}$ is used in the encoder $f_\theta(x, \tilde{p})$ to obtain the instructed representation. All past prompt parameters remain active and shared, enabling dynamic reuse instead of freezing as in static prompt allocation.

To further enhance representation stability, we adopt the same contrastive regularization term as in HiDe-Prompt. Specifically, for each previously encountered class $c \in \mathcal{Y}^{(i)}, i=1, \dots, t-1$, its instructed and uninstructed representations are approximated by class-wise Gaussian distributions $\mathcal{G}_c$ and $\hat{\mathcal{G}}_c$. Let $\tilde{P}_t = \{ \tilde{p}_1, \tilde{p}_2, \dots, \tilde{p}_{N_t} \}$ denote the set of instance-specific prompts dynamically selected by the task-specific router $\mathcal{R}_t$ for all training samples in $\mathcal{D}_t$. We define the current task embeddings as $\mathcal{H}_t = \{ f_\theta(\boldsymbol{x}_i^{(t)}, \tilde{p}_i) \mid i = 1,\dots,N_t \}$ and $\boldsymbol{\mu}_c$ be the mean of $\mathcal{G}_c$. The contrastive loss is computed as:
\begin{align}
\mathcal{L}_{\mathrm{CR}} &= 
    \sum_{h \in \mathcal{H}_t} 
    \sum_{i = 1}^{t - 1} \sum_{c \in \mathcal{Y}^{(i)}}
    \mathrm{log} \left( 
        \frac{\exp(\boldsymbol{h} \cdot \boldsymbol{\mu}_c / \tau)}
        {Z(h)}
    \right),\\
Z(h)&=\sum_{\boldsymbol{h'} \in \mathcal{H}_t} \exp(\boldsymbol{h} \cdot \boldsymbol{h'} / \tau) + 
        \sum_{i = 1}^{t - 1} \sum_{c' \in \mathcal{Y}^{(i)}} \exp(\boldsymbol{h} \cdot \boldsymbol{\mu}_{c'} / \tau)
\end{align}
Here, $Z(h)$ acts as the partition function, collecting all positive and negative pairs for normalization and $\tau$ is the temperature that is set to 0.8. The overall loss for WTP combines classification and contrastive objectives:
\begin{align}\label{eq:wtp_loss}
\mathcal{L}_{\mathrm{WTP}}(\psi, \tilde{P}_t) = \mathcal{L}_{\mathrm{CE}}(\psi, \tilde{P}_t) + \lambda \mathcal{L}_{\mathrm{CR}},
\end{align}
where $\lambda$ is a balancing hyperparameter. Following the WTP stage, we further refine the classifier parameters $\psi$ through a dedicated objective known as task-adaptive prediction (TAP). This stage aims to mitigate classifier bias by accounting for the distributional properties of all previously observed classes. Specifically, TAP optimizes $h_\psi$ using pseudo features sampled from Gaussian approximations of prior class representations. The TAP loss is defined as:
The TAP loss is defined as:
\begin{equation} \label{eq:tap_loss}
\begin{aligned}
    \mathcal{L}_\mathrm{TAP}(\psi) =
    & \sum_{i = 1}^t
      \sum_{c \in \mathcal{Y}^{(i)}}
      \sum_{\boldsymbol{h} \in \mathcal{H}_{i, c}}
      - \log \left(
      \frac{
        \exp\bigl(h_\psi(\boldsymbol{h})[c]\bigr)
      }{
        Z_{\mathrm{TAP}}(\boldsymbol{h})
      }
      \right)
\end{aligned}
\end{equation}
where $Z_{\mathrm{TAP}}(\boldsymbol{h}) = \sum_{j = 1}^t \sum_{c' \in \mathcal{Y}^{(j)}} \exp\bigl(h_\psi(\boldsymbol{h})[c']\bigr)$ is the normalization factor (partition function) over all previously observed classes and $\mathcal{H}_{i, c}$ contains pseudo representations sampled from the Gaussian prototype $\mathcal{G}_c$ of class $c$ from task $\mathcal{T}_i$. This step enhances the alignment of the classifier with the evolving feature space, thus improving robustness against forgetting.

At test time, our method employs a lightweight auxiliary task predictor $\hat{h}_\omega: \mathbb{R}^D \rightarrow \mathbb{R}^T$, trained with the Task-Identity Inference (TII) objective, to infer the task identity from the uninstructed representation $f_\theta(x)$. This module is trained using a cross-entropy loss over pseudo features sampled from the approximate Gaussian distributions $\hat{\mathcal{G}}_c$ of previously encountered classes. Formally, the TII objective is defined as:
\begin{align} \label{eq:tii_loss}
    \mathcal{L}_{\mathrm{TII}}(\omega) = 
    \sum_{c \in \mathcal{Y}_t}
    \sum_{\hat{\boldsymbol{h}} \in \hat{\mathcal{H}}_c}
    - \log \left(
        \frac{
            \exp(\hat{h}_\omega(\hat{\boldsymbol{h}})[c])
        }{
            \sum_{c' \in \mathcal{Y}_t} \exp(\hat{h}_\omega(\hat{\boldsymbol{h}})[c'])
        }
    \right),
\end{align}
where $\hat{\mathcal{H}}_c$ contains pseudo representations sampled from the uninstructed class prototype $\hat{\mathcal{G}}_c$ for $c \in \mathcal{Y}^{(i)}$ and $i = 1, \dots, t$..

During inference, the predicted task index from $\hat{h}_\omega$ determines the corresponding router $\mathcal{R}_t$, which computes prompt relevance scores based on input features and selects the top-$k$ prompts $\mathcal{K}(h)$ from the shared pool. These prompts are then aggregated into a composite vector $\tilde{p}$, which conditions the encoder to produce the final prediction: $\hat{y} = h_\psi(f_\theta(x, \tilde{p}))$. Please refer to Algorithm~\ref{alg:hash} for more details.

\section{Experimental Details}
\subsection{Additional Benchmark Details}
In this section, we provide detailed information on the benchmarks used in our work.

\paragraph{Split CIFAR-100}: CIFAR-100~\cite{krizhevsky2009learning} is a widely used image classification benchmark consisting of images from 100 classes. The dataset contains diverse natural images, including animals, vehicles, and various everyday objects. Each image is annotated with both fine and coarse labels, enabling hierarchical classification tasks. For class-incremental learning settings, the 100 classes are randomly divided into 10 incremental tasks, each comprising a unique set of classes.

\paragraph{Split ImageNet-R}: ImageNet-R~\cite{hendrycks2021many} is a variant of the ImageNet benchmark, containing images from 200 classes that are selected to emphasize robustness to distribution shifts. The dataset includes diverse renditions of objects such as paintings, cartoons, sketches, and other non-photographic styles, covering a broad range of natural and man-made categories. These classes are also randomly divided into 10 distinct incremental tasks.

\paragraph{Split CUB-200}: CUB-200~\cite{wah2011caltech} is a fine-grained image classification dataset containing images from 200 bird species. The dataset features a diverse collection of natural bird photographs, capturing various poses, backgrounds, and environmental conditions. Each image is annotated with a species-level class label, supporting detailed visual recognition research. For class-incremental learning, the 200 categories are randomly split into 10 sequential tasks, each with a distinct subset of bird species.

\paragraph{5-Datasets}: 5-Datasets\cite{ebrahimi2020adversarial} is a composite continual learning dataset that integrates five widely used image classification datasets: CIFAR-10~\cite{krizhevsky2009learning}, MNIST~\cite{lecun1998gradient}, Fashion-MNIST~\cite{xiao2017fashion}, SVHN~\cite{netzer2011reading}, and notMNIST~\cite{bulatov2011notmnist}. Each component dataset presents distinct visual characteristics, ranging from natural images and street view numbers to handwritten and typewritten digits or fashion items. In this benchmark, each dataset is treated as a separate incremental task, enabling the evaluation of methods under substantial distributional shifts and task heterogeneity.

\subsection{Additional Comparison Methods Details}
\begin{table}
\centering
\caption{Performance comparison of \textbf{adapter-based} continual learning methods using ViT-B/16 with Sup-21K weights. Here we present Final Average Accuracy (FAA).}
\small
\setlength{\tabcolsep}{7pt}
\begin{tabular}{lcc}
\toprule
Method & Split CIFAR-100 & Split CUB-200  \\ \midrule
C-ADA               & 88.25           & 76.13          \\
LAE                 & 85.33           & 80.97          \\
ADAM + Adapter             & 87.29           & 85.84          \\
EASE                     & 87.76           & 84.65          \\
InfLoRA                     & 88.31           & 80.78          \\

\textbf{Hash~(Ours)}                      & \textbf{95.02}  & \textbf{91.34} \\ \bottomrule
\end{tabular}
\label{tab:adapter}
\end{table}
\begin{table}
\centering
\caption{Performance comparison of pre-trained \textbf{model-based} continual learning methods using ViT-B/16 with Sup-21K weights. Here we present Final Average Accuracy (FAA).}
\small
\setlength{\tabcolsep}{7pt}
\begin{tabular}{lcc}
\toprule
Method & Split CIFAR-100 & Split CUB-200  \\ \midrule
ADAM + VPT-D               & 85.04           & 85.28          \\
ADAM + SSF                 & 85.27           & 85.67          \\
ADAM + Adapter             & 87.29           & 85.84          \\
RanPAC                     & 92.20           & 90.30          \\

\textbf{Hash~(Ours)}                      & \textbf{95.02}  & \textbf{91.34} \\ \bottomrule
\end{tabular}
\label{tab:ptm}
\end{table}
\begin{table*}[t]
  \centering
  \caption{Overall performance comparison on Split CIFAR-100 and Split ImageNet-R. We present Final Average Accuracy (FAA), Cumulative Average Accuracy (CAA), and Average Forgetting Measure (FM) of all methods under different pre-trained models.}
\small
\setlength{\tabcolsep}{6pt}
\begin{tabular}{llllllll}
\toprule
\multicolumn{1}{c}{\multirow{2}{*}{PTM}} & \multicolumn{1}{c}{\multirow{2}{*}{Method}} & \multicolumn{3}{c}{Split CIFAR-100} & \multicolumn{3}{c}{Split Imagenet-R} \\ \cmidrule(l){3-5} \cmidrule(r){6-8}
\multicolumn{1}{c}{}                     & \multicolumn{1}{c}{}                        & \multicolumn{1}{c}{\textbf{FAA} ($\uparrow$)} & \multicolumn{1}{c}{\textbf{CAA}($\uparrow$)} & \multicolumn{1}{c}{FM($\downarrow$)} & \multicolumn{1}{c}{\textbf{FAA} ($\uparrow$)} & \multicolumn{1}{c}{\textbf{CAA}($\uparrow$)} & \multicolumn{1}{c}{FM($\downarrow$)} \\ 
\midrule

\multirow{8}{*}{iBOT-21K}                  
&L2P         & $79.13 \pm 1.25$ & $85.13 \pm 0.05$ & $7.50 \pm 1.21$ & $61.31 \pm 0.50$ & $68.81 \pm 0.52$ & $10.72 \pm 0.40$ \\
&DualPrompt  & $78.84 \pm 0.47$ & $86.16 \pm 0.02$ & $8.84 \pm 0.67$ & $58.69 \pm 0.61$ & $66.61 \pm 0.67$ & $11.75 \pm 0.92$ \\
&CODA-Prompt & $80.83 \pm 0.27$ & $87.02 \pm 0.20$ & $7.50 \pm 0.25$ & $61.22 \pm 0.35$ & $66.76 \pm 0.37$ & $9.66 \pm 0.20$ \\
&CPrompt    & $82.14 \pm 0.32$ & $88.09 \pm 0.09$ & $7.02 \pm 0.24$ & $74.42 \pm 0.18$ & $79.19 \pm 0.27$ & $7.02 \pm 0.36$ \\
&HiDe-Prompt & $93.02 \pm 0.15$ & $94.56 \pm 0.05$ & $1.26 \pm 0.13$ & $70.83 \pm 0.17$ & $73.23 \pm 0.08$ & $\textbf{6.77} \pm 0.23$ \\
& NoRGa & $94.76 \pm 0.15$ & $95.86 \pm 0.31$ & $1.34 \pm 0.14$ & $73.06 \pm 0.26$ & $77.46 \pm 0.42$ & $6.88 \pm 0.49$ \\ 

& Hash~(Ours) & $\textbf{95.12} \pm 0.24$ & $\textbf{95.97} \pm 0.21$ & $\textbf{1.22} \pm 0.17$ & $\textbf{76.97} \pm 0.18$ & $\textbf{81.87} \pm 0.29$ & $6.84 \pm 0.39$\\ 
\midrule
\multirow{8}{*}{iBOT-1K} 
& L2P        & $75.51 \pm 0.88$ & $82.53 \pm 1.10$ & $6.80 \pm 1.70$ & $59.43 \pm 0.28$ & $66.83 \pm 0.92$ & $11.33 \pm 1.25$ \\
&DualPrompt  & $76.21 \pm 1.00$ & $83.54 \pm 1.23$ & $9.89 \pm 1.81$ & $60.41 \pm 0.76$ & $66.87 \pm 0.41$ & $9.21 \pm 0.43$ \\
&CODA-Prompt & $79.11 \pm 1.02$ & $86.21 \pm 0.49$ & $7.69 \pm 1.57$ & $66.56 \pm 0.68$ & $73.14 \pm 0.57$ & $7.22 \pm 0.38$ \\
&CPrompt    & $83.12 \pm 0.39$ & $89.53 \pm 0.08$ & $6.42 \pm 0.27$ & $72.42 \pm 0.18$ & $75.98 \pm 0.23$ & $7.05 \pm 0.21$ \\
&HiDe-Prompt & $93.48 \pm 0.11$ & $95.02 \pm 0.01$ & $1.63 \pm 0.10$ & $71.33 \pm 0.21$ & $73.62 \pm 0.13$ & $7.11 \pm 0.02$ \\
& NoRGa & $94.01 \pm 0.04$ & $95.11 \pm 0.35$ & $\textbf{1.61} \pm 0.30$ & $72.77 \pm 0.20$ & $76.55 \pm 0.46$ & $7.10 \pm 0.39$ \\ 

& Hash~(Ours) & $\textbf{94.59} \pm 0.07$ & $\textbf{95.98} \pm 0.24$ & $1.72 \pm 0.19$ & $\textbf{75.01} \pm 0.18$ & $\textbf{78.45} \pm 0.22$ & $\textbf{6.99} \pm 0.19$\\ 
\midrule
\multirow{8}{*}{DINO-1K} 
& L2P        & $72.23 \pm 0.35$ & $79.71 \pm 1.26$ & $8.37 \pm 2.30$ & $57.21 \pm 0.69$ & $64.09 \pm 0.74$ & $7.47 \pm 0.96$ \\
&DualPrompt  & $73.95 \pm 0.49$ & $81.85 \pm 0.59$ & $9.32 \pm 1.42$ & $57.98 \pm 0.71$ & $65.39 \pm 0.27$ & $9.32 \pm 0.69$ \\
&CODA-Prompt & $77.50 \pm 0.64$ & $84.81 \pm 0.30$ & $8.10 \pm 0.01$ & $63.15 \pm 0.39$ & $69.73 \pm 0.25$ & $6.86 \pm 0.11$ \\
&CPrompt    & $81.98 \pm 0.52$ & $89.82 \pm 0.43$ & $9.14 \pm 0.66$ & $71.92 \pm 0.38$ & $76.23 \pm 0.32$ & $6.77 \pm 0.64$ \\
&HiDe-Prompt & $92.51 \pm 0.11$ & $94.25 \pm 0.01$ & $1.67 \pm 0.20$ & $68.11 \pm 0.18$ & $71.70 \pm 0.01$ & $6.45 \pm 0.58$ \\
& NoRGa & $93.43 \pm 0.33$ & $94.65 \pm 0.62$ & $1.65 \pm 0.25$ & $71.77 \pm 0.44$ & $75.76 \pm 0.49$ & $6.42 \pm 0.68$ \\ 

& Hash~(Ours) & $\textbf{94.12} \pm 0.09$ & $\textbf{95.21} \pm 0.17$ & $\textbf{1.59} \pm 0.24$ & $\textbf{73.98} \pm 0.23$ & $\textbf{77.74} \pm 0.12$ & $\textbf{6.25} \pm 0.44$\\ 
\midrule
\multirow{8}{*}{MoCo-1K} 
& L2P        & $77.24 \pm 0.69$ & $83.73 \pm 0.70$ & $5.57 \pm 0.75$ & $54.13 \pm 0.67$ & $62.09 \pm 0.76$ & $\textbf{4.88} \pm 0.42$ \\
&DualPrompt  & $77.56 \pm 0.63$ & $84.37 \pm 0.51$ & $6.54 \pm 0.50$ & $54.45 \pm 0.30$ & $62.92 \pm 0.41$ & $5.34 \pm 0.41$ \\
&CODA-Prompt & $77.83 \pm 0.34$ & $84.97 \pm 0.23$ & $12.60 \pm 0.02$ & $55.75 \pm 0.26$ & $65.49 \pm 0.36$ & $10.46 \pm 0.04$ \\
&CPrompt    & $84.12 \pm 0.29$ & $89.23 \pm 0.33$ & $6.92 \pm 0.15$ & $64.79 \pm 0.26$ & $70.64 \pm 0.14$ & $9.73 \pm 0.33$ \\
&HiDe-Prompt & $91.57 \pm 0.20$ & $93.70 \pm 0.01$ & $\textbf{1.51} \pm 0.17$ & $63.77 \pm 0.49$ & $68.26 \pm 0.01$ & $9.37 \pm 0.71$ \\
& NoRGa & $93.52 \pm 0.06$ & $94.94 \pm 0.29$ & $1.63 \pm 0.13$ & $64.52 \pm 0.16$ & $70.21 \pm 0.64$ & $9.06 \pm 0.19$ \\

& Hash~(Ours) & $\textbf{94.37} \pm 0.09$ & $\textbf{95.62} \pm 0.21$ & $1.54 \pm 0.13$ & $\textbf{67.09} \pm 0.13$ & $\textbf{72.12} \pm 0.09$ & $8.14 \pm 0.25$\\ 
\bottomrule
\end{tabular}
\label{tab:cifar_imagenet_all}
\end{table*}
In this section, we provide a detailed description of the methods compared in our work. 

\paragraph{L2P}: L2P~\cite{wang2022learning} is a continual learning framework that leverages a pool of learnable prompts to guide a frozen pre-trained transformer model. During each incremental task, L2P dynamically selects and updates prompts relevant to the current data, enabling the model to adapt to new tasks without revisiting previous data or altering the backbone weights. This approach effectively mitigates catastrophic forgetting and facilitates efficient knowledge integration across tasks.

\paragraph{DualPrompt}: DualPrompt~\cite{wang2022dualprompt} is a prompt-based continual learning method designed for transformer architectures. It introduces two types of prompts: task-shared prompts that capture general knowledge across all tasks, and task-specific prompts that focus on information unique to each task. By jointly optimizing both prompt types, DualPrompt achieves a balance between knowledge retention and task adaptability, significantly improving performance in class-incremental learning scenarios.

\paragraph{CODA-Prompt}: CODA-Prompt~\cite{smith2023coda} introduces a set of learnable prompt components, which are assembled into input-conditioned prompts using an attention-based scheme. Unlike previous methods, all prompting parameters are optimized end-to-end with the task loss, enabling greater capacity and adaptability for rehearsal-free continual learning.

\paragraph{C-Prompt}: CPrompt~\cite{gao2024consistent} addresses the training-testing inconsistency in prompt-based continual learning by introducing classifier consistency learning (CCL) and prompt consistency learning (PCL). CCL exposes prompts to all classifiers during training, while PCL improves prediction robustness and prompt selection using random prompt sampling and a multi-key mechanism. This design achieves more consistent and effective rehearsal-free continual learning.

\paragraph{HidePrompt}: HiDe-Prompt~\cite{wang2023hierarchical} proposes a hierarchical decomposition framework for prompt-based continual learning under self-supervised pretraining. It disentangles the continual learning objective into three components—within-task prediction, task-identity inference, and task-adaptive prediction—and optimizes them jointly using task-specific prompts and structured contrastive regularization. This approach enhances robustness to pretraining paradigms and improves generalization in task-incremental scenarios.

\paragraph{NoRGa}: NoRGa~\cite{le2024mixture} is a prompt-based continual learning framework that reinterprets prefix tuning as adding task-specific experts to a pre-trained mixture-of-experts model. By embedding non-linear residual gating into the prefix scoring mechanism, NoRGa improves parameter estimation and sample efficiency while preserving model compactness and mitigating catastrophic forgetting.

\begin{figure*}[ht]
\begin{center}
\centerline{\includegraphics[width=\textwidth]{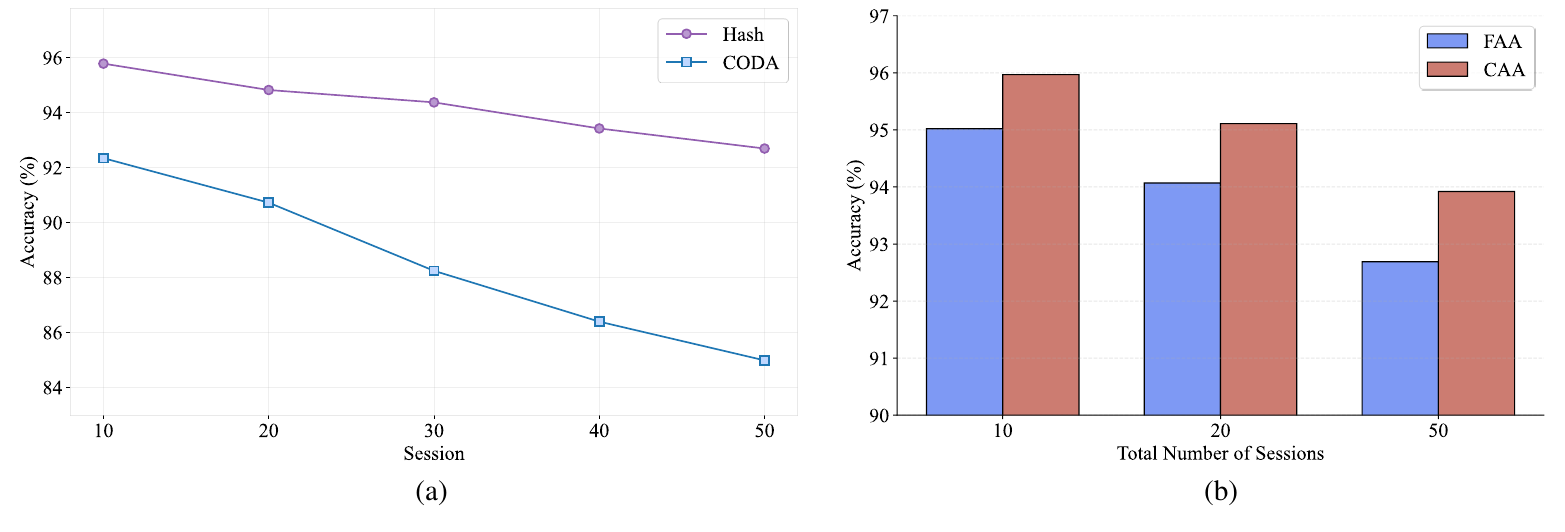}}
\caption{Long sequence incremental analysis: (a) Test accuracy after every 10 sessions in the 50-session setting on Split CIFAR-100, comparing Hash and CODA. (b) Test accuracy across different total session numbers.}
\label{fig:session}
\end{center}
\end{figure*}

\subsection{Evaluation Metric}
\textbf{Evaluation Metrics.} We adopt three standard metrics to assess continual learning performance: Final Average Accuracy (FAA), Cumulative Average Accuracy (CAA), and Forgetting Measure (FM). Let $a_{i, t}$ denote the accuracy on task $\mathcal{T}_i$ after learning task $\mathcal{T}_t$, and define the average accuracy after learning $t$ tasks as $A_t = \frac{1}{t} \sum_{i=1}^{t} a_{i, t}$. Then, FAA is computed as $A_T = \frac{1}{T} \sum_{i=1}^{T} a_{i, T}$, CAA is given by $\frac{1}{T} \sum_{t=1}^{T} A_t = \frac{1}{T} \sum_{t=1}^{T} \left( \frac{1}{t} \sum_{i=1}^{t} a_{i, t} \right)$, and FM is calculated as $\frac{1}{T - 1} \sum_{i=1}^{T - 1} \max_{1 \le t < T} \left( a_{i, t} - a_{i, T} \right)$. FAA is the primary metric to evaluate final performance, CAA reflects performance across the entire learning process, and FM quantifies the extent of forgetting.

\subsection{Implementation Details}
For each setting, we report the average performance over three independent runs using different random seeds to account for training variance. In more detail, L2P~\cite{wang2022learning} is configured with a total of $N=30$ prompts, a prompt length of $L_p=5$, and Top-$K$ key selection with $K=5$. For DualPrompt~\cite{wang2022dualprompt}, we use a prompt length of $L_g=5$ for the task-shared prompts $\boldsymbol{g}$, which are inserted into layers 1 and 2, and a prompt length of $L_e=20$ for the task-specific prompts $\boldsymbol{e}$, inserted at layers 3 through 5. S-Prompt++~\cite{wang2022s} follows a similar configuration to DualPrompt, but replaces all task-shared prompts with task-specific ones; specifically, the task-specific prompts are inserted into layers 1-5 with a prompt length of $L_e=20$. For CODA-Prompt~\cite{smith2023coda}, we use $N=100$ prompts, each of length $L_p=8$, inserted into layers 1-5. HiDe-Prompt~\cite{wang2023hierarchical} and NoRGa~\cite{le2024mixture} shares the overall architectural design of S-Prompt++, yet differs by substituting the task-specific keys with an auxiliary classifier $\hat{h}_{\omega}$ for task identification. Hash modifies the HiDe-Prompt architecture by adopting a mixture-of-experts framework, where the Top-$K$ experts with $K=2$ are selected for each input. We note that our reported results are based on inference with batch size 1. We have verified that performance remains consistent across different inference batch sizes, but omit these additional configurations for brevity.

\begin{figure*}[ht]
\begin{center}
\centerline{\includegraphics[width=\textwidth]{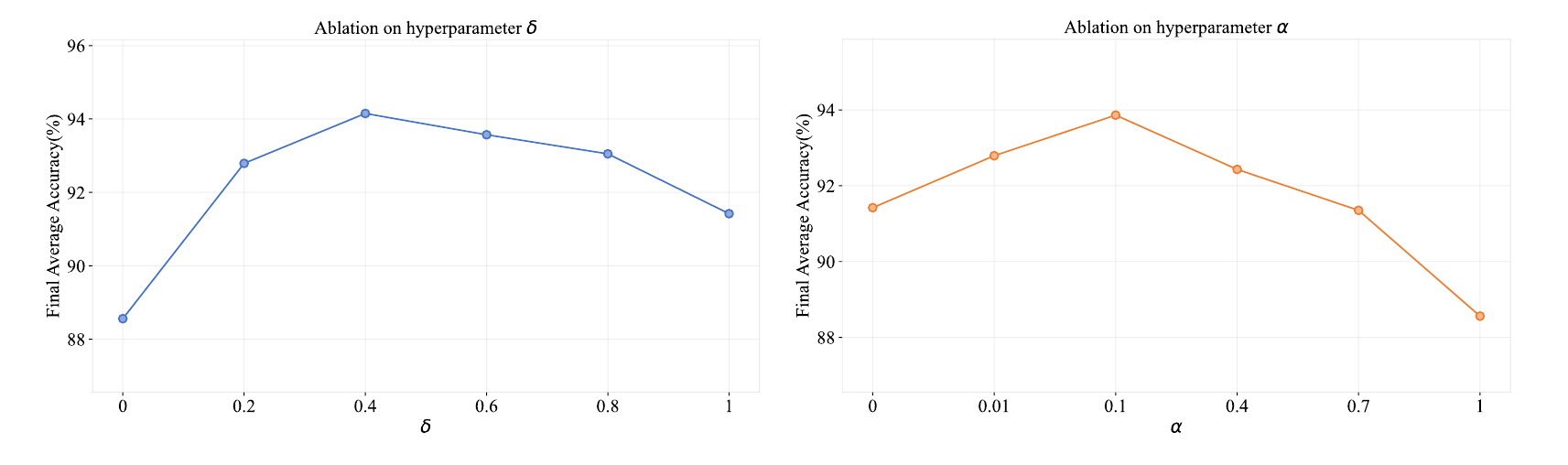}}
\caption{Ablation study on the hyperparameter of history-aware module.}
\label{fig:hyper_abl}
\end{center}
\end{figure*}

\section{Extended Results}
\subsection{Long Sequence Incremental Analysis}
We also present results under the challenging 50-session setting. As shown in Figure~\ref{fig:session}, our method maintains strong performance even with 50 sessions, exhibiting only a minor drop in accuracy compared to the 20-session setting. Notably, the performance degradation is slower than other prompt-sharing methods, highlighting the anti-forgetting advantage of our history-aware module. 

\subsection{Comparison with Adapter-based Methods}
We compare our method with several recent adapter-based continual learning approaches. LAE~\cite{gao2023unified} and ADAM~\cite{zhou2025revisiting} propose unified frameworks for parameter-efficient tuning (PET) via model ensembling and modular adaptation. C-ADA~\cite{gao2024beyond} introduces Continual Adapter Layers with a Scale-and-Shift module to enable task-specific attention and feature transformation. EASE~\cite{zhou2024expandable} achieves rehearsal-free continual learning by iteratively adapting multiple adapter modules through repeated forward propagation, while InfLoRA~\cite{liang2024inflora} leverages low-rank adaptation with an online class-specific memory to stabilize task transitions.

As shown in Table~\ref{tab:adapter}, our method achieves a Final Average Accuracy (FAA) of 95.02\% on Split CIFAR-100, surpassing the next best method, InfLoRA, by 6.71\%. On the fine-grained Split CUB-200 benchmark, our approach attains 91.34\% FAA, 5.5\% higher than ADAM + Adapter. Adapter-based methods provide parameter efficiency and modularity in continual learning, but often require careful placement. In contrast, our prompt-based method uses dynamic prompt composition and instance-adaptive routing, allowing for more flexible and precise knowledge sharing and task adaptation.. 

\subsection{Comparison with Pre-trained Model-based Methods}
Recent research has demonstrated that continual learning can greatly benefit from pre-trained models (PTMs), especially when paired with parameter-efficient tuning (PEFT) strategies. Methods such as ADAM~\cite{zhou2025revisiting} and RanPAC~\cite{mcdonnell2023ranpac} typically adapt the backbone model only during the first task, employing techniques like FiLM or prompt tuning. By leveraging strong pre-trained representations, these approaches often attain high accuracy on the initial task sequence.

However, these methods exhibit notable limitations when presented with new tasks, as the backbone remains fixed after the initial adaptation. This constraint can hinder the model's ability to acquire and disentangle features specific to novel tasks, particularly in the presence of significant distribution shift. 

To address this, we propose a method that maintains a global prompt pool and employs a dynamic routing mechanism to select task-relevant prompts during both training and inference. As shown in Table~\ref{tab:ptm}, our approach consistently outperforms first-task adaptation baselines on both Split CIFAR-100 and Split CUB-200 benchmarks. Specifically, we achieve a Final Average Accuracy (FAA) of 95.02\% on CIFAR-100, exceeding ADAM and RanPAC by 2.82\%, with similar improvements observed on CUB-200 (91.34\% FAA). These results highlight the effectiveness of continual prompt adaptation in overcoming the rigidity of static backbone tuning.

\begin{table}[ht]
  \centering
  \caption{Effect of different history-aware dynamic routing strategies.}
  \small
  \setlength{\tabcolsep}{6pt}
	\begin{tabular}{lcccc}
    \toprule
    \multirow{2}{*}{HDR ($\psi$)} & \multicolumn{2}{c}{Split CIFAR-100} & \multicolumn{2}{c}{Split ImageNet-R} \\
    \cmidrule(l){2-3} \cmidrule(r){4-5}
     & FAA & CAA & FAA & CAA \\ \midrule
    None         &88.56 &90.12 &75.92 &79.53 \\
    Logarithmic  &91.02 &93.21 &76.84 &80.91 \\
    Polynomial   &93.12 &94.38 &76.95 &81.05 \\
    \textbf{Stepwise~(Ours)} &\textbf{94.15} &\textbf{95.03} &\textbf{77.03} &\textbf{81.21} \\
    \bottomrule
    \end{tabular}
    \label{tab:dr}
\end{table}
\begin{table}[ht]
  \centering
  \caption{Effect of different history-aware gradient modulation strategies.}
  \small
  \setlength{\tabcolsep}{6pt}
	\begin{tabular}{lcccc}
	  \toprule
        \multirow{2}{*}{HGM ($\gamma$)} & \multicolumn{2}{c}{Split CIFAR-100} & \multicolumn{2}{c}{Split ImageNet-R} \\
        \cmidrule(l){2-3} \cmidrule(r){4-5}
         & FAA & CAA & FAA & CAA \\ \midrule
        None         &88.56 &90.12 &75.92 &79.53 \\
        Inverse      &92.45 &93.88 &76.23 &80.24 \\
        Exponential  &93.24 &94.19 &76.37 &80.45 \\
        \textbf{Piecewise~(Ours)} &\textbf{93.86} &\textbf{94.77} &\textbf{76.49} &\textbf{80.82} \\
        \bottomrule
	\end{tabular}
    \label{tab:gm}
\end{table}

\subsection{Results under Alternative Pre-training Paradigms}
To complement the main paper, which reports results based on supervised ImageNet-21K (Sup-21K) pre-training, we provide additional experimental results for all compared methods under alternative pre-training paradigms, including iBOT \cite{zhouimage}, DINO \cite{caron2021emerging}, and MoCo \cite{chen2021empirical}. These results serve to evaluate the robustness and generalizability of various prompt-based continual learning approaches across diverse pre-training strategies. All experiments are conducted under the same experimental settings as described in the main paper, ensuring fair and consistent comparisons.

Across all alternative pre-training paradigms, our method consistently achieves high accuracy and maintains low forgetting rates. This demonstrates that the shared prompt and history-aware routing strategies not only ensure robust performance under different pre-trained backbones, but also offer strong scalability to diverse continual learning scenarios.

\section{Additional Ablation Results}
\subsection{Ablation on History-Aware Modulator}
We study several alternative penalty and modulation functions to validate the design choices in our history-aware routing and gradient modulation components.

\paragraph{Dynamic Routing.} We compare three variants: (i) Logarithmic penalty $\psi(h)=\log(1+h)$, which applies a smooth and conservative decay that favors stable reuse of frequently activated prompts; (ii) Polynomial penalty $\psi(h)=h^\gamma$, where $\gamma>1$, enforcing stronger suppression on heavily used prompts to encourage diversity; and (iii) our adopted Stepwise penalty, which subtracts a constant value $\delta$ for top-$k$ prompts, offering a practical trade-off between simplicity and balancing. Logarithmic tends to under-penalize, while Polynomial may over-penalize and hurt performance. Stepwise yields better load balancing without sacrificing reuse.

\paragraph{Gradient Modulation.} We examine: (i) Inverse Scaling $\gamma(h)=1/(1+\beta h)$, offering gradual decay and flexible control; (ii) Exponential Decay $\gamma(h)=\exp(-\beta h)$, enabling stronger suppression but more sensitive to $\beta$; and (iii) our adopted Piecewise Constant strategy, where only top-$k$ experts are modulated with a constant factor $\alpha<1$. Compared to continuous modulation, our approach achieves better balance between stability and plasticity while being easier to tune.

As illustrated in Figure~\ref{fig:hyper_abl}, we select the optimal hyperparameters $\delta$ = 0.4 and $\alpha$ = 0.1 based on the ablation study. Empirical results (Table~\ref{tab:dr} and Table~\ref{tab:gm}) are consistent with our analysis: our piecewise and stepwise strategies achieve superior performance and stability across datasets, validating their effectiveness and ease of tuning.

\begin{table}[t]
\centering
\caption{Comparison of training times for NoGRa and Hash. All experiments were conducted on a single NVIDIA A100 GPU.}
\small
\setlength{\tabcolsep}{4pt}
\begin{tabular}{lcccc}
\toprule
Method & CIFAR-100 & ImageNet-R & CUB-200 & 5-Datasets \\ \midrule
Hide & 2.80h & 2.67h & 1.04h & 24.06h \\
NoRGa & 2.85h & 2.70h & 1.10h & 24.23h \\ 
Hash & 2.66h & 2.49h & 0.96h & 23.52h \\
\bottomrule
\end{tabular}
\label{tab:time}
\end{table}


\section{Training Cost Analysis}
All experiments are conducted on a single A100 GPU, and the corresponding training times are reported in Table~\ref{tab:time}. Compared to HiDe-Prompt and NoRGa, our approach not only requires less training time but also achieves superior performance across all evaluated benchmarks. This computational efficiency stems from our architectural design choices: we employ a smaller total number of prompts through shared pooling, use shorter prompt lengths per instance due to sparse top-k selection, and inject prompts into fewer layers compared to Hide-Prompt. These design decisions substantially reduce the computational overhead during both forward and backward passes, while our dynamic routing mechanism ensures that the reduced capacity is allocated more effectively. As a result, our method offers a more efficient optimization process without sacrificing accuracy, demonstrating the advantages of parameter sharing and selective activation in achieving both computational efficiency and strong continual learning performance.


\end{document}